# Converting Depth Images and Point Clouds for Feature-based Pose Estimation

Robert Lösch[1], Mark Sastuba[2], Jonas Toth[1], and Bernhard Jung[1]

*Abstract*— In recent years, depth sensors have become more and more affordable and have found their way into a growing amount of robotic systems. However, mono- or multi-modal sensor registration, often a necessary step for further processing, faces many challenges on raw depth images or point clouds. This paper presents a method of converting depth data into images capable of visualizing spatial details that are basically hidden in traditional depth images. After noise removal, a neighborhood of points forms two normal vectors whose difference is encoded into this new conversion. Compared to Bearing Angle images, our method yields brighter, higher-contrast images with more visible contours and more details. We tested feature-based pose estimation of both conversions in a visual odometry task and RGB-D SLAM. For all tested features, AKAZE, ORB, SIFT, and SURF, our new Flexion images yield better results than Bearing Angle images and show great potential to bridge the gap between depth data and classical computer vision. Source code is available here: https://rlsch.github.io/depth-flexion-conversion.

## I. INTRODUCTION AND RELATED WORK

Depth sensors such as LiDAR, Time-of-Flight (ToF) or stereo camera systems are getting more affordable and are therefore integrated into an increasing number of robotic systems and everyday technology. This consequently extends to the use of depth data, both point clouds and depth/range images, in a variety of traditional robotic algorithms like visual odometry (VO). In order to do so, point clouds or range/depth images need to be aligned to each other. Beside odometry, Simultaneous Localization and Mapping (SLAM) [1], (global) localization [2], [3], and place recognition [4] are also relevant fields of application. Creating realistic virtual models requires accurate mapping of visual information about the environment onto range information [5]. This can only be achieved if cameras and depth sensors are extrinsically calibrated. With point correspondences, the calibration becomes a camera pose estimation problem. However, range images in general are lacking point features [5]. Point cloud and depth image registration (*e.g.*, using Iterative Closest Points (ICP) [6]) face different challenges like noise and outliers, only partial overlap, density difference, and scale variation between sensors [7]. One possible solution is to process range images and convert them into a new representation that highlights significantly more details of the environment than depth images, like the Bearing Angle (BA) image.

[1]Robert Lösch, Bernhard Jung are, Jonas Toth was with Institute of Computer Science, Technical University Bergakademie Freiberg, Germany {Robert.Loesch,jung}@informatik.tu-freiberg.de, development@jonas-toth.eu

[2]Mark Sastuba is with German Centre for Rail Traffic Research at the Federal Railway Authority, Germany sastubam@dzsf.bund.de

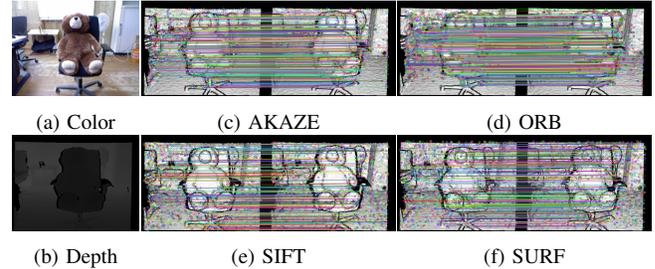

(a) Color  (c) AKAZE  (d) ORB
(b) Depth  (e) SIFT  (f) SURF

Fig. 1. Depth images (*e.g.*, Fig. 1b) in general are a poor geometrical representation of the environment and they usually lack distinguishable point features. Our proposed Flexion image (*e.g.*, Figs. 1c to 1f) converts depth images into a more descriptive format and enables them to be used in traditional computer vision algorithms like feature matching. Depicted are two consecutive images of *fr1/teddy* of TUM RGB-D data set [8].

Harati *et al.* [9] first introduced BA images in order to segment planar surfaces from range images as features for 3D indoor SLAM. They argue that raw point cloud data is too redundant to be used in mapping and instead use BA images to characterize surfaces in a specified direction. The authors propose region- and edge-based methods, both utilizing BA images. They criticize the use of (median) filters for noise reduction when using surface normals. However, they also suggest noise removal in a later stage. For image segmentation, they choose horizontal and vertical BA images and later argue that these two directions suffice for successful segmentation. The edge-based method applies edge detection on each BA image separately and a final image is obtained by combining the results with a logical OR.

In the same year, Scaramuzza *et al.* [5] introduced BA images for extrinsically calibrating a camera and a 3D laser range finder. Instead of introducing landmarks that can be seen by both camera and laser, they want to use image features as point correspondences between camera image and laser point cloud. In doing so, this allows to solve the problem by using standard camera pose estimation algorithms. In order to identify point correspondences in the range data, the authors enhance it by highlighting discontinuities and orientation changes along specific directions with the help of a slightly differently defined BA. Their calibration procedure comprises three steps: converting the range image into a BA image (in all four directions, but assuming only the horizontal direction is necessary), manually selecting at least four point correspondences between the camera image and BA image, and achieving extrinsic calibration through a camera pose estimation algorithm and non-linear refinement.

Lin *et al.* [10] use BA images for registration of point clouds. They convert them into diagonal BA images, extract SURF [11] features, and match image pairs. The optimal rotation matrix is then determined by using a least squares approximation with 3D points from the top half of the best corresponding pairs. The point clouds can be aligned based on the determined relative transformation matrix between the image pair. Since the proposed algorithm does not require iterative processing, it reduces computation time significantly and is ten times faster than generalized-ICP [12].

Due to bad lighting conditions, Zhuang *et al.* [13] perform scene recognition based on 3D laser data and BA images instead of color images. After converting 3D range data into diagonal BA images, they not only extract local SIFT [14] features from them but also extract a global spatial feature from 3D laser scanning data. In order to do so, they use maximal 2D coverage area in the 3D scan. After conducting experiments, the authors confirm BA images' superior performance in representing detailed scene information.

This paper introduces a novel method called Flexion image, inspired by BA images, for converting depth data into a detail-highlighting image representation (Fig. 1). Unlike BA images, the resulting Flexion image is more descriptive and rotation-invariant, based on the relation between surface normals calculated from horizontal/vertical and diagonal/antidiagonal vectors. The remaining sections of this paper are organized as follows: Sec. II discusses preliminary considerations to be taken into account before Sec. III recapitulates BA images and defines the creation of Flexion images, Sec. IV compares the performance of these methods by quantitatively comparing the results of VO of a synthetic scene and the results of RGB-D SLAM of that synthetic scene and of real world data, and Sec. V concludes the paper.

## II. PRELIMINARY CONSIDERATIONS

Usually, as seen in the literature, BA images are calculated based on point cloud data. However, RGB-D images combined with the intrinsics of the sensor are also suitable for these types of depth conversions. For that reason and since we want to demonstrate that Flexion images can be used to register structured point clouds as well, this section elaborates on some exemplary sensor models of (depth) cameras and laser scanners. Probably the most common and most used camera model is the pinhole camera model usually applied for ordinary RGB or even RGB-D cameras. By contrast, laser scanners work differently and their principle is based on angular measures. Hence, equirectangular images are a natural choice for depicting laser scanner data from terrestrial laser scanners (TLSs). In the following, both models are briefly explained.

### A. Pinhole camera model

For the sake of simplicity, we assume a simple distortion-free pinhole camera model with the intrinsic parameters focal lengths $f_x, f_y$, principal point $(c_x, c_y)$ and skew coefficient $s$. The projection is given by

$$\begin{pmatrix}u\\v\\1\end{pmatrix} = \begin{bmatrix}f_x & s & c_x\\0 & f_y & c_y\\0 & 0 & 1\end{bmatrix}\begin{pmatrix}X\\Y\\Z\end{pmatrix}\frac{1}{Z}, \quad (1)$$

which maps a 3D point $\mathbf{P} = (X, Y, Z)^\top$ in camera frame to its corresponding 2D pixel position $(u, v)^\top$ on the image sensor. The back-projection describes the inverse transformation such that

$$\begin{pmatrix}X\\Y\\Z\end{pmatrix} = \begin{bmatrix}\frac{1}{f_x} & \frac{-s}{f_x f_y} & \frac{c_y s - c_x f_y}{f_x f_y}\\0 & \frac{1}{f_y} & \frac{-c_y}{f_y}\\0 & 0 & 1\end{bmatrix}\begin{pmatrix}u\\v\\1\end{pmatrix}d \quad (2)$$

by using the (orthographic) depth value $d = Z$, which is stored in a corresponding depth map (or depth image).

### B. Spherical camera model

Omnidirectional cameras and TLSs are defined by the angular measuring resolution in azimuth and polar direction. The equirectangular projection is a natural choice for depicting range and depth data as 2D images. The azimuth angle $\varphi \in [0, 2\pi)$ is mapped onto the width $w$ and the polar angle $\theta \in [0, \pi]$ is mapped onto the height $h$ of the image. Depending on the image size, the horizontal angular resolution $\Delta\varphi$ and vertical one $\Delta\theta$ of the image are:

$$\Delta\varphi = \frac{\varphi_{max} - \varphi_{min}}{w}, \quad \Delta\theta = \frac{\theta_{max} - \theta_{min}}{h} \quad (3)$$

where the horizontal Field of view (FoV) is defined as $[\varphi_{min}, \varphi_{max}]$ and the vertical is defined as $[\theta_{min}, \theta_{max}]$.

Let $\mathbf{P} = (X, Y, Z)^\top$ be a 3D point in camera and laser frame, respectively. The conversion to spherical coordinates is given by:

$$r = \sqrt{X^2 + Y^2 + Z^2} \quad (4)$$

$$\varphi = \arctan2\left(\frac{Y}{X}\right), \quad \theta = \arccos\left(\frac{Z}{r}\right). \quad (5)$$

The pixel values then are simply converted by

$$\begin{pmatrix}u\\v\end{pmatrix} = \begin{pmatrix}\frac{\varphi}{\Delta\varphi}\\\frac{\theta}{\Delta\theta}\end{pmatrix}. \quad (6)$$

The back-projection describes the inverse transformation such that:

$$\varphi = \varphi_{min} + u\Delta\varphi, \quad \theta = \theta_{min} + v\Delta\theta \quad (7)$$

$$\begin{pmatrix}X\\Y\\Z\end{pmatrix} = \begin{pmatrix}\sin(\theta)\cos(\varphi)\\\sin(\theta)\sin(\varphi)\\\cos(\theta)\end{pmatrix}r, \quad (8)$$

by using the range value $r$, which is stored in a corresponding range image.

## III. RANGE DATA CONVERSION

All types of densely structured depth data (point clouds, range/depth images) can be processed in order to produce detail-highlighting images of the captured environment. In order to prevent ambiguities regarding BA image calculation, we decide on a definition and discuss their peculiarities. Then, we define the calculation of Flexion images and compare them to BA images based on their definitions.

## A. Bearing Angle image

Harati *et al.* [9] and Scaramuzza *et al.* [5] introduce Bearing Angle images in two slightly different forms at about the same time. In general, BA images can be created by assigning each pixel the angle between the laser beam of a point and the line connecting it to its neighboring point (Fig. 2). That point can be located on a horizontal, vertical, or on one of both diagonal lines. For a horizontal BA image, the neighboring point could be either to the left [9] or to the right [5] of the point in question. Going forward, we adopt the definition of Harati *et al.* [9].

*1) Calculation:* Let sensor center $\mathbf{C}$, Point $\mathbf{P}_{i,j}$, and $\mathbf{P}_{i-1,j}$ be an arbitrary triangle (Fig. 2). While previous definitions [5], [9] of the BA are based on the cosine theorem, for consistency reasons, we calculate it directly between vectors $\overrightarrow{\mathbf{P}_{i,j}}$ and $\overrightarrow{\mathbf{P}_{i-1,j} - \mathbf{P}_{i,j}}$ yielding

$$\beta_{i,j} = \arccos\left(\frac{(\mathbf{P}_{i,j})^\top (\mathbf{P}_{i,j} - \mathbf{P}_{i-1,j})}{\|\mathbf{P}_{i,j}\|_2 \|\mathbf{P}_{i,j} - \mathbf{P}_{i-1,j}\|_2}\right). \quad (9)$$

The BA value needs to be mapped to a color map. For feature extraction, a grayscale image with a color depth of $b = 8$ bit should suffice. With the Bearing Angle being in the range $\beta \in (0, \pi)$ rad, the grayscale value $g$ for a color depth of $b$ can be calculated with

$$g_{i,j} = \left\lfloor (2^b - 1) \frac{\beta_{i,j}}{\pi} \right\rfloor. \quad (10)$$

*2) Characteristics:* The definition of the Bearing Angle in combination with the operating principle of a depth sensor results in limited rotation and viewpoint invariance. Since a calculation of BA images depends on a predefined pixel relationship, a roll rotation of more than $45°$ is equivalent to a change of this relation. A horizontal BA image can become a rotated vertical BA image (compare Fig. 3).

As Fig. 4 demonstrates, a rotation in pitch or yaw (depending on the type of BA image) as well as a translation of the camera center have an effect on BA images. The larger the distance of a point to the camera center on a flat surface, the flatter the laser beams impinge on the surface and, therefore, the Bearing Angle becomes larger or smaller. This effect also becomes apparent in a gradient shading of flat surfaces along the calculation direction of BA images (usually the flat ground, see Fig. 3).

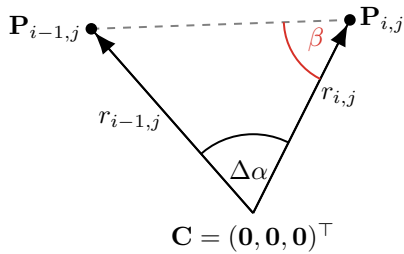

Fig. 2. Schematic drawing of two light rays and corresponding Bearing Angle $\beta$. Points $\mathbf{P}_{i-1,j}$ and $\mathbf{P}_{i,j}$ are two adjacent points of a point cloud with a distance of $r_{i-1,j}$ and $r_{i,j}$ from the sensor center $\mathbf{C}$ respectively and an angular resolution of $\Delta\alpha$.

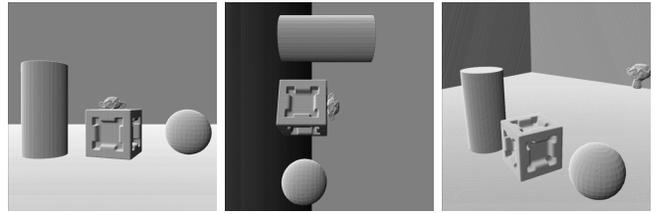

Fig. 3. Bearing Angle images of a synthetic scene. Due to the nature of its definition, the Bearing Angle is not invariant to rotation and viewpoint changes. The depth images were converted with the diagonal (top left to bottom right) point relation of the Bearing Angle formula.

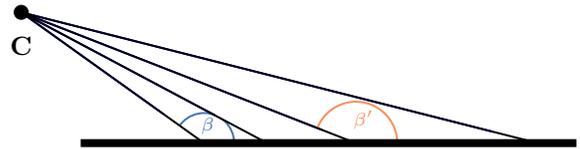

Fig. 4. Different incidence angles on a flat surface cause Bearing Angles to differ, resulting in gradient shading of the surface in the image.

## B. Flexion image

Instead of using the Bearing Angle to optically improve laser scans or depth images, Flexion images are based on surface curvature. Classical methods of obtaining a curvature measure yield neither a robust nor a high-contrast result (see Fig. 5). Another method of encoding surface characteristics are surface normals [15], [16]. Since noise can have a considerable effect on the direction of normals, a pre-processing step smoothes the data before the Flexion image is created.

*1) Pre-processing:* When processing depth data, geometric features like edges or corners are usually most interesting. These sharp changes need to stay unaltered by a filter. The widely used median filter [17] is edge-preserving and reduces salt-and-pepper noise effectively. The filter convolutes an image, replacing each pixel with the median of its neighborhood, for example an $n \times m$ "window" with $n, m \in \mathbb{N}_{2k+1}$. No floating point operations are required and the filter can be implemented with $O(n)$ [18] time complexity.

*2) Calculation:* The basic idea of Flexion images is to estimate surface curvature by determining two surface normals based on their own set of neighboring pixels and calculating their difference. As Fig. 6 illustrates, for each point $\mathbf{P}_{i,j}$, the first normal $\vec{n}_{1,i,j}$ is based on its horizontal and vertical neighbors (Eq. (11)) and the second normal $\vec{n}_{2,i,j}$ is based on its diagonal and antidiagonal neighbors (Eq. (12)).

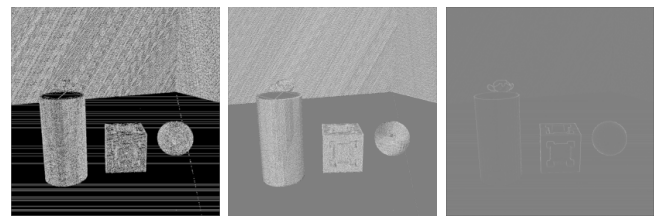

(a) Gaussian curvature    (b) Mean curvature    (c) Max curvature

Fig. 5. Pixel-wise calculation of Gaussian (Fig. 5a), Mean (Fig. 5b), or Max (Fig. 5c) curvature yield noisy and low-contrast images.

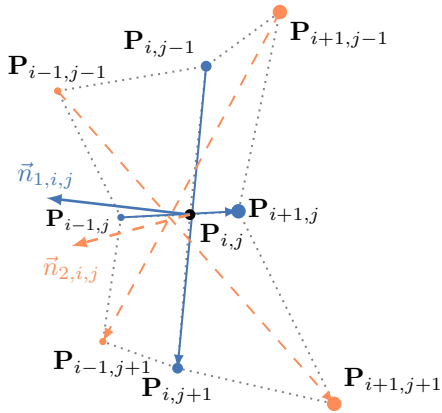

Fig. 6. Graphical representation of the Flexion image calculation. Horizontal, vertical (blue) and diagonal, antidiagonal (orange) neighbors of $\mathbf{P}_{i,j}$ that form the two normal vectors defining the Flexion image. Due to the curvature of the surface, horizontal/vertical neighbors define a different normal vector ($\vec{n}_{1,i,j}$, blue) than diagonal/antidiagonal neighbors ($\vec{n}_{2,i,j}$, orange, dashed). Thereby, the local shape of the surface influences the angle between both normal vectors that is proportional to the Flexion $\mathcal{F}_{i,j}$.

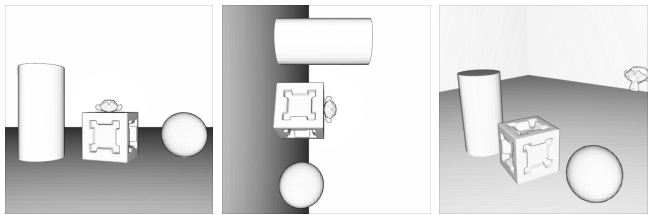

Fig. 7. Flexion images of a synthetic scene. Their appearance is very plastic and the shading effects give a good sense for depth. The conversion is rotation invariant. Depicted are the same camera positions as in Fig. 3.

$$\vec{n}_{1,i,j} = \frac{\mathbf{P}_{i,j-1} - \mathbf{P}_{i,j+1}}{\|\mathbf{P}_{i,j-1} - \mathbf{P}_{i,j+1}\|_2} \times \frac{\mathbf{P}_{i-1,j} - \mathbf{P}_{i+1,j}}{\|\mathbf{P}_{i-1,j} - \mathbf{P}_{i+1,j}\|_2} \quad (11)$$

$$\vec{n}_{2,i,j} = \frac{\mathbf{P}_{i-1,j-1} - \mathbf{P}_{i+1,j+1}}{\|\mathbf{P}_{i-1,j-1} - \mathbf{P}_{i+1,j+1}\|_2} \times \frac{\mathbf{P}_{i-1,j+1} - \mathbf{P}_{i+1,j-1}}{\|\mathbf{P}_{i-1,j+1} - \mathbf{P}_{i+1,j-1}\|_2} \quad (12)$$

These two normals usually differ and span an angle (Fig. 6). It is to be noted that both normals $\vec{n}_{1,i,j}$ and $\vec{n}_{2,i,j}$ are not unit length but each factor of their cross product is. Finally, the Flexion $\mathcal{F}_{i,j}$ at point $\mathbf{P}_{i,j}$ is defined as

$$\mathcal{F}_{i,j} = \left| (\vec{n}_{1,i,j})^\top \vec{n}_{2,i,j} \right|, \quad (13)$$

with $|\cdot|$ being the absolute value. Since the length of the normals are $\|\vec{n}_{1,i,j}\|_2, \|\vec{n}_{2,i,j}\|_2 \in [0,1]$, the value of $\mathcal{F}_{i,j}$ is bound to $\mathcal{F}_{i,j} \in [0,1]$. This value is then linearly scaled to a grayscale value analogous to Eq. (10).

*3) Characteristics:* Contrary to BA images, Flexion images use all neighboring points for calculation which makes a Flexion image rotation invariant. A roll rotation of more than $45°$ only swaps $\vec{n}_{1,i,j}$ and $\vec{n}_{2,i,j}$ which does not change the final value of $\mathcal{F}$. Fig. 7 demonstrates rotation invariance, which becomes obvious when comparing to Fig. 3.

Similar to BA images, a flat surface that is approximately perpendicular to the sensor plane has an almost constant shading. Flat surfaces at a sharp angle (like the ground plane) show gradient shading. This is due to perspective transformation which leads to shorter normals and therefore darker shades of gray, as Fig. 8 explains. In general, the more an angle, spanned by two vectors, departs from a right angle, the shorter the length of the resulting cross-product vector will be. Since normals $\vec{n}_{1,i,j}$ and $\vec{n}_{2,i,j}$ are not normalized, $\mathcal{F}$, the result of a scalar product, inherits this attribute.

*4) Variations:* Due to its simple form, the definition of Flexion images can easily be modified. In order to decrease the susceptibility to noise, the neighboring points after next, *e.g.*, $\mathbf{P}_{i\pm 2,j\pm 2}$, can be used to form normal vectors. Because of the $3\times 3$ neighborhood as described in Eqs. (11) and (12), it could be termed Flexion$_{3\times 3}$, in general Flexion$_{n\times n}$ with $n$ being the size of the grid used to form normal vectors. Increasing the distance to point $\mathbf{P}_{i,j}$ has two effects: smoothing the image and widening edges. Fig. 9 illustrates these effects. Other variations might be normalizing $\vec{n}_{1,i,j}$ and $\vec{n}_{2,i,j}$ or, similar to BA images, calculating the angle between them. In order to map the angle to an grayscale image and yield similar results in terms of color, we calculate:

$$\mathcal{F}_{\text{angle } i,j} = 1.0 - \frac{1}{\pi} \arccos\left( \frac{(\vec{n}_{1,i,j})^\top \vec{n}_{2,i,j}}{\|\vec{n}_{1,i,j}\|_2 \|\vec{n}_{2,i,j}\|_2} \right). \quad (14)$$

Figs. 9g and 9h show results of the two aforementioned variations of the Flexion image creation. Due to the normalization of the vectors, the gradient shading effect described in Fig. 8 does not occur anymore.

## IV. COMPARISON OF FLEXION AND BA IMAGES

In order to cover a widespread assessment of Flexion and BA images, the first part evaluates results of a simple VO with multiple feature descriptors and ORB-SLAM3 [19] as state-of-the-art reference of the "'Multi-FoV' synthetic dataset" [20]. The second part aims at evaluating ORB-SLAM3 [19] results of parts of the real-world TUM RGB-D data set [8].

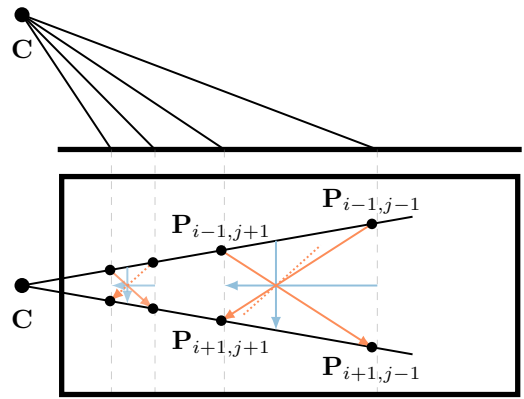

Fig. 8. Side and top view of laser beams hitting a flat surface. The angle between the diagonals (orange) decreases with increasing distance to sensor center $\mathbf{C}$ which results in shorter normals. This yields smaller Flexion values and creates a gradient shading of flat surfaces. The dotted diagonal is elongated and parallel shifted to the right to make the change visible.

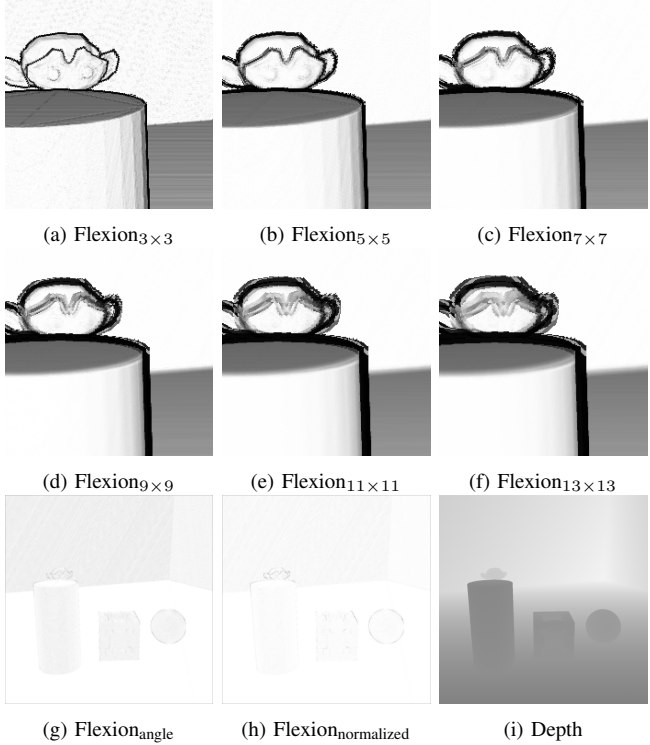

Fig. 9. Variants of Flexion images. Using a larger distance to calculate Flexion values results in less noise and wider edges. Figs. 9b to 9f show less noise, smoother surfaces, and thicker edges respectively. Also, finer details, *e.g.*, in the monkey head, are lost. Figs. 9g and 9h lack gradient shading of the ground plane and show, especially in a synthetic scene, little detail.

## A. Method and Metrics

Benchmark for VO or SLAM performance is the respective result obtained with color images (*e.g.*, Fig. 10a) and the ground truth trajectory. The evaluation compares Flexion$_{n \times n}$ images with $n \in [3, 5, 7, 9, 11, 13]$ (*e.g.*, Fig. 10c), Flexion$_{angle}$, Flexion$_{normalized}$, and all four BA image variations (*e.g.*, Fig. 10d). All converted images are mapped to 8 bit grayscale and the evaluation is based on absolute trajectory error (ATE) [8] and relative pose error (RPE) [8]. While ATE measures global consistency, RPE measures translational drift. We use RMSE for ATE and mean translational error for RPE, *i.e.*, the default implementation of [8]. In the following tables, the best three results of every column are bold, the best value is underlined as well. Tabs. III and V to VIII present the median of eleven ORB-SLAM3 runs.

## B. "'Multi-FoV' synthetic dataset"

Following a path through a synthetic city scene (482.59 meters long), the data set [20] provides color images (Fig. 10a), depth data (Fig. 10b), ground truth, and intrinsics.

*1) Visual Odometry:* The simple odometry algorithm uses the default implementation of OpenCV's (v4.5.5) [21] AKAZE [22], ORB [23], SIFT [14], SURF [11], and the default functions *knnMatch* (with Lowe's ratio test [14]), *findEssentialMat* (with RANSAC [24]), and *recoverPose*.

Tab. I shows ATE and Tab. II shows RPE for all tested features and image variations. AKAZE and ORB work best for depth based images. SIFT yields best color performance but only works well for Flexion$_{3-9}$. In contrast, SURF has the most problems with depth based images, also yields worst color performance, except for Flexion$_{7,9}$ that even top color images. Considering Flexion images only, ATE and RPE tend to get higher with larger $n$. Therefore, Flexion$_{11,13}$ mostly yield highest values, sometimes significantly larger than Flexion$_{3-9}$. Flexion$_{angle}$ and Flexion$_{normalized}$ also yield usually bad results, except for ORB. BA images peak with ORB and partially with AKAZE. In general, Flexion$_{3-9}$ yield better odometry results than BA but similar when using ORB. It is also noteworthy that color images are not always the best performing images and are not always among the top three results. A reason might be that this synthetic scene has no noise in depth data and possibly repetitive textures. The usually bad performance of Flexion$_{angle}$ and Flexion$_{normalized}$ might be explained by their low-contrast, low-detail images.

TABLE I
ATE [m] OF VISUAL ODOMETRY ON SYNTHETIC SCENE

| Variant | AKAZE | ORB | SIFT | SURF |
|---|---|---|---|---|
| Flexion$_{3\times3}$ | **3.99** | **5.20** | 7.65 | 32.03 |
| Flexion$_{5\times5}$ | 7.87 | **5.70** | **6.45** | 30.47 |
| Flexion$_{7\times7}$ | 8.70 | 13.75 | 9.22 | **5.10** |
| Flexion$_{9\times9}$ | **7.13** | 6.72 | **7.33** | **6.10** |
| Flexion$_{11\times11}$ | 14.07 | 11.41 | 44.20 | 45.33 |
| Flexion$_{13\times13}$ | 8.66 | 16.00 | 53.87 | 41.48 |
| Flexion$_{angle}$ | 40.56 | 11.05 | 40.49 | 31.04 |
| Flexion$_{normalized}$ | 19.29 | 7.63 | | 36.74 |
| BA horizontal | 22.53 | 13.36 | 38.87 | 44.23 |
| BA vertical | 11.01 | 7.44 | 42.53 | 35.16 |
| BA diagonal | 11.87 | 8.00 | 44.13 | 55.67 |
| BA antidiagonal | 16.72 | 14.70 | 31.28 | 27.60 |
| Color | 6.19 | **4.89** | **3.58** | 12.24 |

*2) RGB-D SLAM:* Using the converted depth images as color input for ORB-SLAM3 [19] yields the results presented

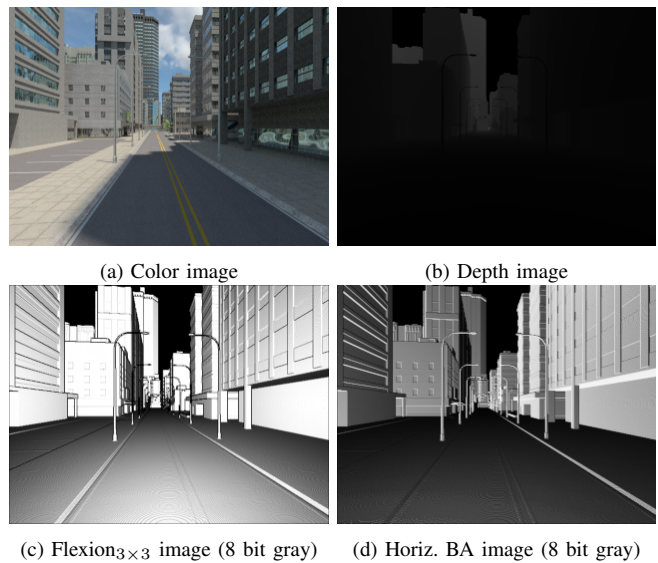

Fig. 10. The first image of the "'Multi-FoV' synthetic dataset" [20] in different implementations. Figs. 10a and 10b belong to the original data set. Figs. 10c and 10d are conversions of depth image Fig. 10b.

## TABLE II
RPE [m] OF VISUAL ODOMETRY ON SYNTHETIC SCENE

| Variant | AKAZE | ORB | SIFT | SURF |
|---|---|---|---|---|
| Flexion$_{3\times3}$ | **47.22** | 55.85 | 54.74 | 87.19 |
| Flexion$_{5\times5}$ | **50.45** | 58.56 | **46.52** | 67.76 |
| Flexion$_{7\times7}$ | 60.51 | 57.69 | 56.19 | **49.47** |
| Flexion$_{9\times9}$ | 56.08 | 54.84 | **52.55** | 51.91 |
| Flexion$_{11\times11}$ | 84.75 | 83.25 | 98.54 | 60.87 |
| Flexion$_{13\times13}$ | 76.05 | 87.64 | 110.16 | 61.05 |
| Flexion$_{angle}$ | 91.58 | 50.47 | 68.22 | 82.45 |
| Flexion$_{normalized}$ | 69.06 | 54.65 | | 69.46 |
| BA horizontal | 69.87 | **49.62** | 92.06 | 69.77 |
| BA vertical | 57.97 | **50.05** | 71.96 | 80.24 |
| BA diagonal | 68.97 | 55.55 | 76.97 | 104.52 |
| BA antidiagonal | **55.91** | **48.38** | 75.10 | 79.71 |
| Color | 56.11 | 53.03 | 51.36 | 56.95 |

## TABLE III
RESULTS OF RGB-D SLAM ON SYNTHETIC SCENE

| Variant | ATE [m] | RPE [m] |
|---|---|---|
| Flexion$_{3\times3}$ | 33.64 | 40.27 |
| Flexion$_{5\times5}$ | 25.53 | 31.24 |
| Flexion$_{7\times7}$ | 12.74 | 15.79 |
| Flexion$_{9\times9}$ | 7.55 | **8.67** |
| Flexion$_{11\times11}$ | **7.14** | **8.84** |
| Flexion$_{13\times13}$ | **7.16** | 8.86 |
| Flexion$_{angle}$ | 39.95 | 51.00 |
| Flexion$_{normalized}$ | 19.01 | 22.26 |
| BA horizontal | 41.37 | 54.17 |
| BA vertical | 27.00 | 32.27 |
| BA diagonal | 30.11 | 35.49 |
| BA antidiagonal | 31.39 | 37.01 |
| Color | **5.25** | **6.97** |

in Tab. III. Probably due to noise, a larger $n$ in Flexion images has a positive effect on ATE and RPE. Flexion$_{7-13}$ yield better or similar results and are, except for color, the best performing images. In opposition, the remaining Flexion variants yield worse results compared to the VO on synthetic data. BA images got a worse ATE and a better RPE than before. While the results were similar to Flexion before, with ORB-SLAM3 the performance is notably worse. This might be due to stricter constraints of the SLAM algorithm.

### C. TUM RGB-D data set

Flexion and BA images were tested with part of the TUM RGB-D data set [8] in four smaller and six larger scenarios. Tab. IV shows ground truth trajectory length for all used scenes. Tabs. V and VII show ATE and Tabs. VI and VIII show RPE of the small and large scenes respectively.

Flexion$_{angle}$ and Flexion$_{normalized}$ achieved such bad results that ORB-SLAM3 often failed to create a map and therefore these algorithms are omitted below. The first two scenes share the same desk setup. Scene *fr1/xyz* yields small errors among all image types. A more complex trajectory on the same desk (*fr1/desk*) shows that Flexion images beat not only BA but also color images. Much clutter might lead to wrong matches in color images but provide much spatial details that benefit Flexion images. Scene *fr1/floor* has only little structure and therefore proves to be the most challenging for depth based images. In *fr1/room*, only Flexion$_{5,7}$ yield comparable results to BA images. Over all, Flexion$_{5,7}$ yield the best performance among all depth based images.

## TABLE IV
GROUND TRUTH TRAJECTORY LENGTH [m] OF TUM RGB-D [8]

| Scene | fr1/xyz | fr1/desk | fr1/floor | fr1/room |
|---|---|---|---|---|
| Length | 7.112 | 9.263 | 12.569 | 15.989 |

| fr2/large_with_loop | fr3/long_office_househ | fr2/pioneer_360 | slam | slam2 | slam3 |
|---|---|---|---|---|---|
| 39.111 | 21.455 | 16.118 | 40.380 | 21.735 | 18.135 |

## TABLE V
ATE [cm] OF RGB-D SLAM ON TUM RGB-D 1/2

| Variant | fr1/xyz | fr1/desk | fr1/floor | fr1/room |
|---|---|---|---|---|
| Flexion$_{3\times3}$ | 3.82 | 28.41 | 83.16 | 130.89 |
| Flexion$_{5\times5}$ | 3.15 | 21.25 | **64.96** | 69.27 |
| Flexion$_{7\times7}$ | 3.11 | 14.25 | **66.26** | **62.37** |
| Flexion$_{9\times9}$ | 3.15 | **9.18** | 75.19 | 122.43 |
| Flexion$_{11\times11}$ | 3.16 | **8.82** | 74.13 | 129.73 |
| Flexion$_{13\times13}$ | 3.11 | **9.53** | 75.99 | 106.49 |
| BA horizontal | **2.64** | 67.64 | 75.68 | 87.96 |
| BA vertical | 2.84 | 67.78 | 74.76 | **69.06** |
| BA diagonal | 3.02 | 61.91 | 86.48 | 73.59 |
| BA antidiagonal | **2.73** | 54.45 | 83.55 | 70.75 |
| Color | **1.04** | 40.68 | **4.70** | **7.30** |

In general, Flexion images perform better than BA images in the smaller scenes whereas results in the larger scenes are more balanced. Scenes *fr2/large_with_loop* and *fr3/long_office_houshold* yield relatively good results. In the first scene, Flexion$_{3-5}$ yield best results, comparable with three out of four BA variants. In the second scene, Flexion performs better than BA and minimally peaks with $n = 7$. BA performance is worse and more unstable. Scenes *fr2/pioneer_360*, *slam2*, and *slam3* yield the worst results (compared to their ground truth trajectory length) with *slam2* being the most challenging scene as even color images yield their worst result. With *360*, Flexion peaks at $n = 9$ with BA being worse, *slam* shows roughly identical results of all depth based methods, *slam2* results are in a similar range with BA being marginally better than Flexion, and *slam3* produces the worst results with Flexion$_{5-9}$ being better than BA.

## V. CONCLUSIONS AND FUTURE WORK

This paper presents a novel method to convert depth data into a geometrically more descriptive format than depth images. While Bearing Angle images are ambiguously defined and lack large rotation invariance, our Flexion images eliminate these drawbacks and prove better performance regarding visual odometry and RGB-D SLAM. The visual

## TABLE VI
RPE [cm] OF RGB-D SLAM ON TUM RGB-D 1/2

| Variant | fr1/xyz | fr1/desk | fr1/floor | fr1/room |
|---|---|---|---|---|
| Flexion$_{3\times3}$ | 4.87 | 37.05 | 148.76 | 152.11 |
| Flexion$_{5\times5}$ | 4.00 | 28.35 | 129.13 | 98.90 |
| Flexion$_{7\times7}$ | 3.98 | 19.39 | 124.78 | **85.81** |
| Flexion$_{9\times9}$ | 4.02 | **13.87** | 122.34 | 149.60 |
| Flexion$_{11\times11}$ | 4.03 | **12.98** | 120.53 | 156.52 |
| Flexion$_{13\times13}$ | 3.98 | **14.29** | 122.23 | 120.58 |
| BA horizontal | **3.43** | 78.70 | 136.50 | 124.39 |
| BA vertical | 3.67 | 78.56 | 122.76 | 93.94 |
| BA diagonal | 3.77 | 72.10 | 129.49 | 101.77 |
| BA antidiagonal | **3.55** | 66.29 | **120.89** | 90.50 |
| Color | **1.45** | 48.55 | **6.24** | **11.31** |

TABLE VII
ATE [$cm$] OF RGB-D SLAM ON TUM RGB-D 2/2

| Variant | | fr2/large_with_loop | fr3/long_office_househ | fr2/pioneer_ | | | |
|---|---|---|---|---|---|---|---|
| | | | | 360 | slam | slam2 | slam3 |
| Flexion | $3\times3$ | **112.51** | 77.78 | 182.14 | 193.81 | 215.72 | 191.96 |
| | $5\times5$ | **112.74** | 64.65 | 173.42 | **187.00** | 215.59 | 181.59 |
| | $7\times7$ | 138.47 | 57.36 | 159.36 | 189.14 | 215.86 | 180.19 |
| | $9\times9$ | 135.98 | 61.35 | 155.12 | 187.69 | 209.96 | 173.70 |
| | $11\times11$ | 136.35 | 61.68 | 156.04 | 191.79 | 210.77 | 175.66 |
| | $13\times13$ | 136.45 | 67.29 | 158.24 | **184.22** | 207.38 | 162.06 |
| BA | horiz. | 113.68 | 94.87 | 191.93 | 192.27 | 217.31 | 195.90 |
| | vert. | 112.87 | 80.48 | – | 194.37 | **205.25** | 192.58 |
| | diag. | 112.87 | 192.08 | 199.31 | 192.49 | 208.10 | 190.98 |
| | antid. | 142.11 | 72.79 | 196.08 | 190.69 | 212.55 | 192.80 |
| Color | | 18.47 | 12.37 | 12.25 | 137.84 | 156.62 | 111.26 |

TABLE VIII
RPE [$cm$] OF RGB-D SLAM ON TUM RGB-D 2/2

| Variant | | fr2/large_with_loop | fr3/long_office_househ | fr2/pioneer_ | | | |
|---|---|---|---|---|---|---|---|
| | | | | 360 | slam | slam2 | slam3 |
| Flexion | $3\times3$ | 141.11 | 93.42 | 221.44 | 243.68 | 276.94 | 259.86 |
| | $5\times5$ | 143.49 | 75.81 | 211.24 | 239.85 | 284.03 | **239.41** |
| | $7\times7$ | 211.79 | 65.91 | 195.12 | **236.84** | 282.42 | 241.34 |
| | $9\times9$ | 251.07 | **70.41** | 186.37 | 242.51 | 282.05 | 246.34 |
| | $11\times11$ | 254.59 | 71.86 | 187.42 | 243.04 | 285.05 | 252.73 |
| | $13\times13$ | 159.37 | 74.04 | 193.47 | 243.54 | 281.85 | 265.85 |
| BA | horiz. | 141.68 | 122.34 | 239.75 | 239.23 | 274.45 | 274.12 |
| | vert. | **137.60** | 92.36 | – | 248.65 | **269.44** | 252.70 |
| | diag. | **138.87** | 188.75 | 251.55 | 239.08 | 266.47 | 250.39 |
| | antid. | 168.77 | 87.81 | 246.73 | **238.23** | 277.49 | 263.00 |
| Color | | 28.56 | 17.15 | 19.86 | 112.87 | 213.78 | 83.88 |

odometry showed that SIFT and SURF perform mediocre while AKAZE and ORB allow for better and partly color image-like performance. ORB-SLAM3 on parts of the TUM RGB-D data set reveals that, depending on the unevenness and size of the environment, Flexion$_{7\times7}$ or Flexion$_{9\times9}$ might be a good alternative when color images are not available. Furthermore, Flexion images seem to work best in smaller areas since spatial features are more visible and therefore contribute more to the level of detail in the image.

The results are promising and Flexion images might be the foundation to exploit range/depth image's full potential. In order for the proposed method to bridge the gap between point clouds and classical computer vision, further evaluation and research towards (global) localization, SLAM, and for example Structure from Motion (SfM) need to be conducted. Further variations of image creation could also be investigated, *e.g.*, including center point $\mathbf{P}_{i,j}$ into calculation. Since classical ICP relies on establishing explicit point correspondences, point-cloud-registration might also be a field Flexion images could explore.

ACKNOWLEDGMENT

This paper is based on a master's thesis by Jonas Toth who not only thought about improving BA images but also analyzed feature performance on BA images, Flexion images, and images created by Gaussian, mean, and max curvature. This activity has received funding from the European Institute of Innovation and Technology (EIT), a body of the European Union, under the Horizon 2020, the EU Framework Programme for Research and Innovation [17019].